\documentclass[journal]{IEEEtran}

\usepackage{cite}

\usepackage{multirow}
\ifCLASSINFOpdf
\usepackage{amssymb}
\else

\fi
\usepackage{amsfonts}
\usepackage{amsmath}
\usepackage{url}
\usepackage{booktabs}
\usepackage{graphicx}

\usepackage{orcidlink}
\usepackage{graphicx}
\usepackage{academicons}
\usepackage{color}

\title{SSF-Net: Spatial-Spectral Fusion Network with Spectral Angle Awareness for Hyperspectral Object Tracking}

\author{Hanzheng~Wang\orcidlink{0000-0001-7276-3216},
        Wei~Li\orcidlink{0000-0001-7015-7335}, \textit{Senior Member, IEEE,}
        Xiang-Gen Xia\orcidlink{0000-0002-5599-7683}, \textit{Fellow, IEEE,}
        Qian~Du\orcidlink{0000-0001-8354-7500}, \textit{Fellow, IEEE,}
        and Jing~Tian\orcidlink{0000-0001-7362-0620}, \textit{Member, IEEE}

\thanks{This work is supported by NSFC Projects of International Cooperation and Exchanges under Grant W2411055. (Corresponding author: Wei Li.)

Hanzheng Wang, Wei Li, and Jing Tian are with the School of Information and Electronics, Beijing Institute of Technology, Beijing 100081, China and the National Key Laboratory of Science and Technology on Space-Born Intelligent Information Processing, Beijing 100081, China and also with  Beijing Institute of Technology, Zhuhai, Guangdong 519088, China (e-mail: hzwangc@bit.edu.cn; liwei089@ieee.org; tianjing@bit.edu.cn).

Xiang-Gen Xia is with the Department of Electrical and Computer Engineering, University of Delaware, Newark, DE 19716, USA (e-mail: xxia@ee.udel.edu).

Qian Du is with the Department of Electrical and Computer Engineering, Mississippi State University, Starkville, MS 39762, USA (e-mail: du@ece.msstate.edu).}}

\begin{document}


\maketitle
%

\begin{abstract}
Hyperspectral video (HSV) offers valuable spatial, spectral, and temporal information simultaneously, making it highly suitable for handling challenges such as background clutter and visual similarity in object tracking. However, existing methods primarily focus on band regrouping and rely on RGB trackers for feature extraction, resulting in limited exploration of spectral information and difficulties in achieving complementary representations of object features. In this paper, a spatial-spectral fusion network with spectral angle awareness (SSF-Net) is proposed for hyperspectral (HS) object tracking. Firstly, to address the issue of insufficient spectral feature extraction in existing networks, a spatial-spectral feature backbone ($S^2$FB) is designed. With the spatial and spectral extraction branch, a joint representation of texture and spectrum is obtained. Secondly, a spectral attention fusion module (SAFM) is presented to capture the intra- and inter-modality correlation to obtain the fused features from the HS and RGB modalities. It can incorporate the visual information into the HS context to form a robust representation. Thirdly, to ensure a more accurate response to the object position, a spectral angle awareness module (SAAM) is designed to investigate the region-level spectral similarity between the template and search images during the prediction stage. Furthermore, a novel spectral angle awareness loss (SAAL) is developed to offer guidance for the SAAM based on similar regions. Finally, to obtain the robust tracking results, a weighted prediction method is considered to combine the HS and RGB predicted motions of objects to leverage the strengths of each modality. Extensive experiments on the HOTC-2020, HOTC-2024, and BihoT datasets demonstrate the effectiveness of the proposed SSF-Net compared with state-of-the-art trackers. The source code will be available at \href{https://github.com/hzwyhc/SSF-Net}{https://github.com/hzwyhc/hsvt}.

\end{abstract}



\section{Introduction}

\begin{figure} [!htb] 
\includegraphics[width=7cm,height=7.5cm]{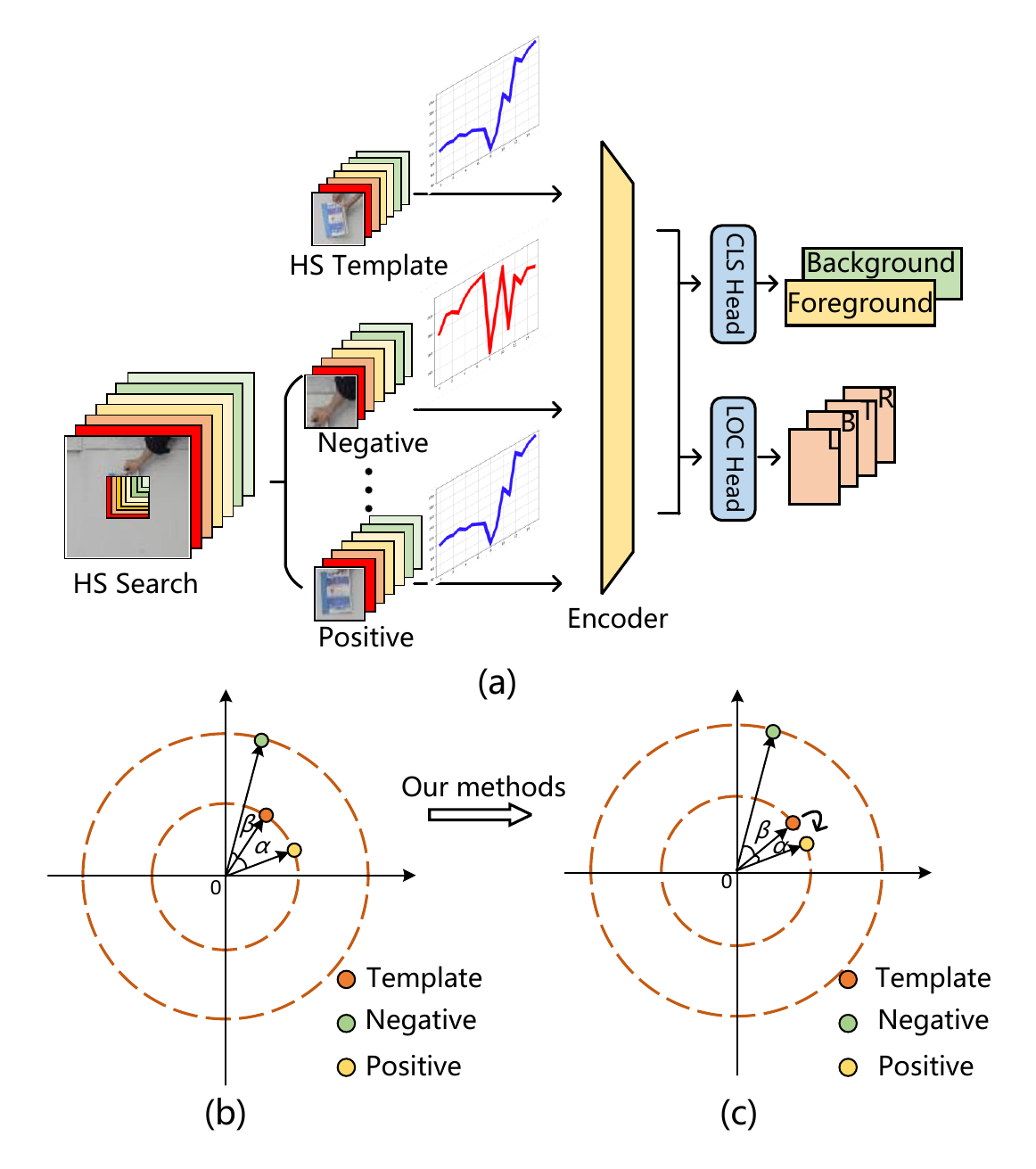}
\centering
\caption{Illustration of the issue in the tracking paradigm and the main idea of our methods.}
\label{fig:1}
\end{figure}

\IEEEPARstart{V}{isual} object tracking has gained significant research attention in recent years due to its applications in autonomous driving \cite{lee2015road, thayalan2023multifocus}, human-computer interaction \cite{chandra2015eye}, and intelligent security \cite{tian2020simultaneous, bi2022multiscale}. It aims to follow the same object in subsequent frames based on its initial location in the first frame of a video sequence. The traditional RGB tracking methods exhibit poor robustness in complex scenarios, such as lighting variations, occlusion, and background interference, making them prone to object drift or tracking failure. Moreover, traditional RGB tracking struggles to effectively distinguish objects from the background when their visual features are similar, such as in camouflaged objects or low-contrast scenes. In contrast, hyperspectral object tracking (HOT) leverages hyperspectral (HS) imaging technology, which captures object information across multiple continuous spectral bands (typically ranging from dozens to hundreds) spanning from visible light to infrared. This enables each pixel to possess rich spectral features, allowing HS data to reveal the physical and chemical composition of objects \cite{he2023spectral}. Consequently, HOT can effectively track objects even in shadows, occlusions, or weak-textured backgrounds. Furthermore, it can utilize spectral-spatial joint modeling, enhancing its adaptability to complex background interference. Therefore, compared to traditional RGB tracking, HOT offers stronger object discrimination and better environmental adaptability, making it a more robust and effective solution in challenging tracking scenarios.

In the earlier studies, hand-crafted features were designed to represent the spectral information of HS data cubes \cite{tang2022target}. Xiong \textit{et al.} \cite{xiong2020material} considered the 3-D spatial-spectral histogram of HS images and constructed an HS object tracking dataset to validate the effectiveness of the proposed methods. Hou \textit{et al.} \cite{hou2022spatial} leveraged the correlation filters to deal with the motion prediction of objects and designed a regularizer for optimization. However, with the advancements in deep learning, hand-crafted features are gradually being replaced by more robust convolutional features that offer improved performance and reliability.

The deep learning-based methods \cite{ouyang2021band, sun2023siamohot} are roughly divided into two categories: band regrouping-based methods and cross-modality fusion-based methods. The former focuses on utilizing the existing RGB trackers to deal with the HS images through transfer learning. Gao \textit{et al.} \cite{gao2023cbff} proposed a feature fusion network, CBFF-Net, to fuse the information from different band groups. Li \textit{et al.} \cite{li2023learning} adopted the auto-encoder learning to capture the intrinsic spectral information for the HS images to be grouped into multiple three-channel false-color images. However, the band regrouping-based methods just utilize the RGB trackers to extract features from the regrouped HS images, which neglects the relationship between spatial-spectral features. There is a lack of feature extraction networks suitable for HS images that consider both spatial and spectral characteristics simultaneously. The latter focuses on combining the advantages of RGB trackers and HS trackers. Liu \textit{et al.} \cite{liu2022siamhyper} designed an RGB and an HS branch to form the robust representation of objects jointly. The RGB and HS images from the same scenario are input to the network to predict the subsequent motions. Zhao \textit{et al.} \cite{zhao2022tftn} proposed a fusion network based on the Transformer encoder, and the robust tracking results are obtained from RGB and HS streams through the self-attention mechanism. Despite these advancements, the impact of various modality fusion methods is not yet fully understood, and there remains a gap in the development of effective feature-level fusion methods for HS object tracking.

Moreover, HS trackers often adopt an RGB tracking framework, which comprises a classification (CLS) head and a localization (LOC) head. However, such an approach does not fully exploit the rich spectral information in HS data, leading to sub-optimal tracking performance. As depicted in Figure \ref{fig:1}(a), the conventional tracking paradigm creates a shared feature space through encoder embeddings. The CLS and LOC heads then predict the class and location of objects, respectively, utilizing depth-wise correlation computations such as generalized Euclidean or cosine metrics—yet they do not account for the spectral correlation between HS template images (TI) and HS search images (SI). As illustrated in Figure \ref{fig:1}(b), for a trained HS tracker, although the classification and location loss converge, the included angles $\alpha$ of the TI and positive SI and $\beta$ of the TI and negative SI are uncertain. The included angles of negative pairs may be larger than positive pairs \cite{ye2020bi}. This problem makes the tracker unable to deal with the disturbance from similar negative objects. Thus, it is necessary to constrain the included angles of both of the above pairs in the common feature space, as shown in Figure \ref{fig:1}(c).

To address the above problems, a spatial-spectral fusion network (SSF-Net) with spectral angle awareness is proposed. Firstly, to extract robust features and facilitate cross-modality feature fusion, a novel spatial-spectral feature backbone $S^2$FB, especially for HS images, is designed, which includes several spatial-spectral convolution blocks ($S^2$CB). Besides, with the downsample layers and residual connections of $S^2$CB, the high-level semantic information is captured, and the subsequent features can be fused naturally. Secondly, to incorporate the visual information from the RGB modality into the HS branch, a spectral attention fusion module (SAFM) is proposed. Compared with other fusion-based trackers, the SAFM captures the cross-spectral context within modalities, and the HS fused features are obtained from the RGB and the HS modalities in a self-adaptive way. Consequently, sufficient visual prompts are supplemented to the HS branch. Finally, to fully exploit the relationships between bands and enhance the robustness of the tracking results, a spectral angle awareness module (SAAM) and a novel spectral angle awareness loss (SAAL) are proposed. Inspired by the spectral angle \cite{kruse1993spectral}, the spectral similarity is determined by the angle between two feature vectors, which can assist the classification branch in location prediction. Between TI and SI, the spectral angle for the regions corresponding to positive samples should be small, whereas for negative samples, it should be large. With the training of the SAAL, the SAAM is capable of accurately predicting the object position, serving as a complement to the CLS head. Moreover, to further improve performance, a weighted prediction module is utilized to integrate predictions from both RGB and HS modalities.

The main contributions of this article are as follows:
\begin{enumerate}
\item A spatial-spectral fusion network with spectral angle awareness (SSF-Net) for HS object tracking is proposed, which integrates the information of the HS and RGB modalities to form the fused feature representation, effectively enhancing the tracking performance.
\item An $S^2$FB is proposed to extract HS features instead of using the off-the-shelf RGB tracker, which aims to model the spatial and spectral context information jointly. Moreover, an SAFM is designed to combine the HS and RGB features with the intra- and inter-modality correlation modeling, thereby generating a more robust HS feature representation.
\item An SAAM is proposed to capture the spectral angle similarity between regions of feature maps to perceive the positions of objects accurately. With the guidance of SAAL, the SAAM can focus on the similar regions between TI and SI. Extensive experiments on public datasets demonstrate the effectiveness of our proposed methods.
\end{enumerate}

The remaining paper is organized as follows. Section II reviews the related works, including RGB tracker, HS tracker, and fusion tracking. Section III introduces the proposed SSF-Net. Extensive ablation and comparative experimental results are shown in Section IV to show the effectiveness of the proposed methods on the HOTC-2020, HOTC-2024, and BihoT \cite{wang2024bihot} datasets. Finally, Section V concludes the article.

\section{Related work}

\subsection{Visual Object Tracker}
The existing research on RGB trackers is roughly divided into two categories: trackers based on Siamese networks and trackers based on Transformers. The Siamese network was originally proposed by SiamFC \cite{bertinetto2016fully}, which extracts features from query images and template images for similarity matching. After that, many works \cite{xie2022correlation, paul2022robust} focus on extracting discriminative features and improving the robustness of tracking performance. Li \textit{et al.} \cite{li2018high} incorporated the region proposal network into the Siamese tracker, and the tracking problem is considered as a one-shot detection task. Guo \textit{et al.} \cite{guo2021graph} proposed a graph attention sub-network to propagate the object information from TI to SI. With the graph attention mechanism, the object structure and part-level information can be considered. Borsuk \textit{et al.} \cite{borsuk2022fear} proposed a fast, efficient, accurate, and robust Siamese visual tracker through a dual-template representation for object model adaptation. Kang \textit{et al.} \cite{kang2023exploring} designed a lightweight bridge module to incorporate the high-level information of deep features into the shallow large-resolution features, which produces better features for the tracking head. In Ref. \cite{taufique2022siamgauss}, a Siamese region proposal network with a Gaussian head is proposed for object tracking. Chen \textit{et al.} \cite{chen2020siamese} proposed a general framework, SiamBAN, to simplify the complex anchor calculations, which is a unified fully convolutional network that directly predicts the category and location of objects, avoiding the anchor configurations. On the other hand, due to the excellent ability to model the correlation of the self-attention mechanism, the Transformer-based trackers \cite{wang2021transformer,sun2020transtrack} have attracted more attention in recent years. Ye \textit{et al.} \cite{ye2022joint} designed a framework named OSTrack, which can extract the features and model relations of TI and SI in a unified way. Blatter \textit{et al.} \cite{blatter2023efficient} introduced a lightweight exemplar Transformer, which utilizes an attention layer based on a single instance for real-time object tracking. Yan \textit{et al.} \cite{yan2021learning} proposed a Transformer-based tracker, STARK, which considers the global spatial and temporal features simultaneously. Zhu \textit{et al.} \cite{zhu2023visual} proposed a multi-modality tracking paradigm based on visual prompts. With the fine-tuning of the foundation model, the Transformer-based tracker can deal with multiple modalities of images.

\subsection{HS Object Tracker}
HS imaging, with its enhanced spectral information, offers rich spectral features, which have proven to be a powerful modality in addressing the challenge of background clutter in visual object tracking. In some early works, hand-crafted features were adopted to represent objects. Nguyer \textit{et al.} \cite{van2010tracking} introduced a framework that utilized radiative transfer theories to estimate object reflectance. Uzkent \textit{et al.} \cite{uzkent2018tracking} proposed a kernelized correlation filter (KCF) based tracker, which can track aerial vehicles using an adjustable cross-modality HS equipment. With the proposed KCF-in-multiple regions-of-interest (ROIs) approach, a reliable region prediction is obtained. Xiong \textit{et al.} \cite{xiong2020material} proposed the material-based tracking framework to deal with the HS tracking problem. Two feature extractors are used to capture the local spectral and spatial features. Thus, texture and spectral information can be obtained respectively. On the other hand, recent works mainly focus on using convolutional features to represent the object. Li \textit{et al.} \cite{li2020bae} proposed a band regrouping method to generate three-channel false-color images to leverage the RGB tracker to deal with the HS images. Li \textit{et al.} \cite{li2023siambag} adopted the pre-trained RGB tracker to predict raw tracking results and designed an image enhancement module to improve the image quality. Chen \textit{et al.} \cite{chen2023spirit} proposed a spectral awareness module to regroup the bands considering the spectral nonlinear interactions, and the dynamic TI matching strategy generated reliable tracking results. Islam \textit{et al.} \cite{islam2023background} proposed a background-aware band selection method that captures the spatial changes of each band. The most important bands can be selected to generate the false-color images.

\subsection{Fusion Tracking}
Multi-modality fusion tracking, which integrates information from different modalities, has become a prominent area of focus in object tracking. One modality combination that has garnered significant attention is the fusion of RGB and infrared modalities. This approach has proven to be highly effective in enhancing tracking performance, especially in challenging conditions. Zhang \textit{et al.} \cite{zhang2019multi} proposed a discriminative model prediction tracker (DMP), which adopted a multi-level fusion mechanism and presented an object estimation network to improve the performance of RGB-T tracking. Zhang \textit{et al.} \cite{zhang2021jointly} proposed an RGB-T tracker, JMMAC, which considered the appearance and motion cues of objects jointly to generate reliable features. The modality fusion weights are obtained from offline-trained networks, avoiding training costs. Lan \textit{et al.} \cite{lan2019robust} proposed a graph-based label prediction model to capture the relationship of the samples and obtain the importance weights of modalities. With the measure of modality discrimination, the model can predict robust tracking results. As for HS object tracking, the existing work has not yet given sufficient attention to this aspect. Liu \textit{et al.} \cite{liu2022siamhyper} proposed a fusion-based tracker SiamHYPER, which extracted both RGB and HS features. With the channel attention mechanism \cite{wang2020eca}, the fused features are obtained. In the inference stage, the RGB and HS branches are used to predict the motion of objects. Zhao \textit{et al.} \cite{zhao2022tftn} proposed a Transformer-based network named TFTN, which captured the cross-modality and intra-modality information of different modalities.


\section{Proposed Method}

\subsection{Framework Overview}

\begin{figure*}
    \centering
    \includegraphics[width=15.0cm,height=10.0cm]{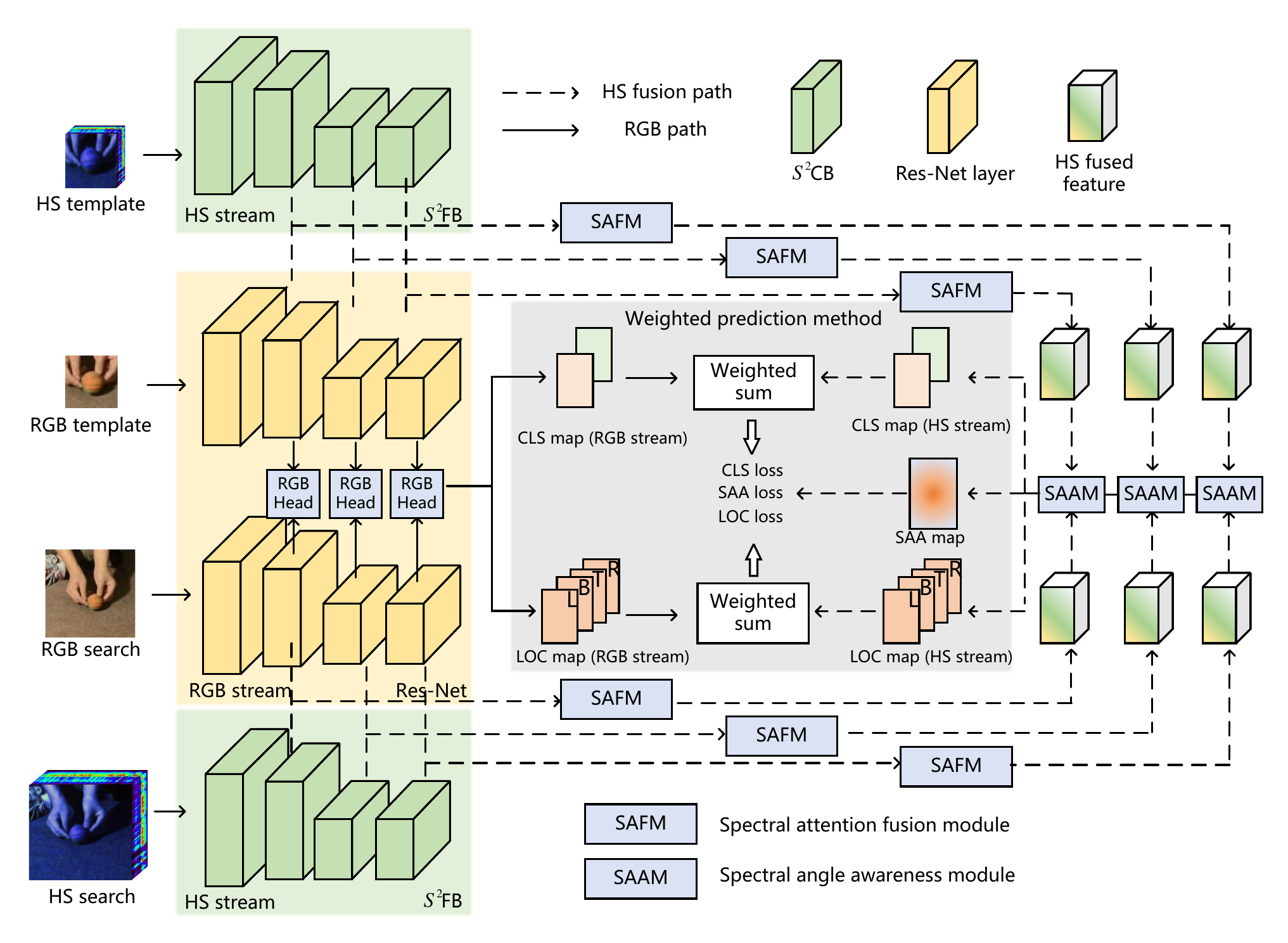}
    \caption{Illustration of the overall structure of our proposed SSF-Net, including the bi-stream feature extraction backbone, spectral attention fusion module, spectral angle awareness module, and the weighted prediction method.}
    \label{fig:2}
\end{figure*}

The overall network structure is shown in Figure \ref{fig:2}. Firstly, an RGB stream and an HS stream are used to extract features from the input image. To address the issue of insufficient modeling of HS images using existing methods, $S^2$FB is proposed for feature extraction of HS streams. The $S^2$FB includes the spatial-spectral convolutional block ($S^2$CB), which effectively captures both spatial and spectral information, resulting in more discriminative features for cross-modality fusion tracking. The RGB stream adopts the conventional Res-Net architecture and serves as a backbone network. It is worth noting that the two streams of the SSF-Net backbone network also have four stages. It can simultaneously extract features from different semantic levels, and it facilitates feature fusion with the RGB-stream backbone network at the same stage. In addition, an SAFM is proposed for spectral feature fusion of the RGB-stream and the HS-stream. Specifically, considering that different bands (spectra) contain different semantic information, a spectral attention network is proposed to enhance cross-modality feature representation, which flows from the backbone network to SAFM in a gating mechanism and then performs cross-modality feature fusion to obtain HS fused features. Then, for robust tracking, the object motions on both RGB and HS streams are simultaneously predicted, and a weighted ensemble strategy is adopted to fuse the tracking results of the two streams. The RGB stream utilizes the tracking head of the existing object tracking framework \cite{chen2020siamese}, thus the classification and location maps can be obtained. For the HS stream, the SAAM is proposed, which consists of a spectral angle awareness branch, along with classification and localization branches. The mixed template features and search features from RGB and HS backbones through SAFM are then fed into SAAM for object position prediction. Inspired by spectral angle mapping \cite{kruse1993spectral}, this module calculates the similarity between spectra by measuring the angle between spectral feature vectors. By comparing spectral angles between different regions, the tracker can identify similar regions between two frames. Finally, to effectively train the network, the SAAL is proposed to increase the predicted spectral similarity between the TI and object regions of SI and lead the tracker to focus on the similar regions between TI and SI. After that, the SAAL is combined with classification loss and location loss for multi-task training of the network, which ensures that the network learns to effectively utilize spectral information for accurate tracking.

\subsection{Bi-stream Feature Extraction Backbone}

\begin{figure*}
    \centering
    \includegraphics[width=14.0cm,height=8.5cm]{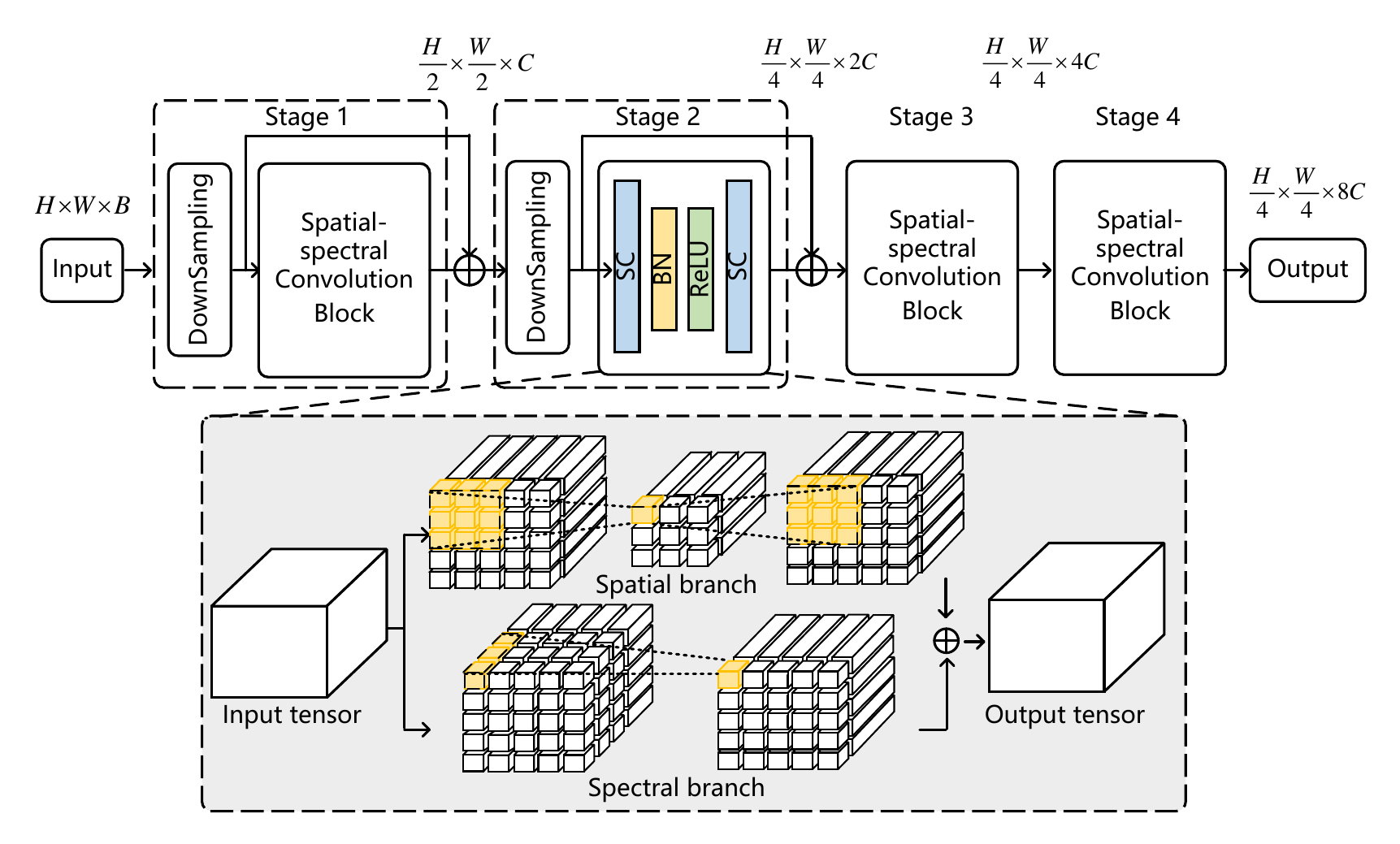}
    \caption{Illustration of our proposed $S^2$FB. It consists of several $S^2$CB with the residual connection. Each $S^2$CB contains $S^2$C, a Batch Normalization layer, and ReLU layers, where $S^2$C consists of parallel spatial feature extraction branches and spectral feature extraction branches.}
    \label{fig:3}
\end{figure*}

To extract the robust features, a bi-stream feature extraction backbone is proposed, which consists of the HS stream and the RGB stream. Most existing HS tracking methods utilize the off-the-shelf models in the RGB domain, such as VGG, ResNet, etc, to extract the HS features. However, an HS image is a data cube enriched with information across numerous spectral bands, which can offer a wealth of multi-dimensional information, and the backbone network pre-trained from three-channel RGB images is insufficient to extract robust spatial-spectral features. 

Thus, we propose a feature extraction backbone $S^2$FB for HS-stream. Its main structure is shown in Figure \ref{fig:3}. The size of input images is $H \times W \times B$, where $H$ and $W$ denote the height and width of the input images, respectively, and $B$ denotes the number of spectral bands. The $S^2$FB consists of four feature extraction stages. Each stage captures different feature responses to the input data. In the early stages, the network focuses on capturing details and local features, while in the later stages, as the network is deepened, it captures higher-level semantic information. Thus, $S^2$FB can extract discriminative features across multiple layers, ranging from shallow to deep. The first two stages, which are referred to as “Stage 1” and “Stage 2”, include $S^2$CB and downsampling layers. The objective of these stages is to ensure that the feature size is compatible with the RGB stream for feature fusion and to capture texture information. The output resolutions of these stages are $\frac{H}{2} \times \frac{W}{2} \times C$ and $\frac{H}{4} \times \frac{W}{4} \times 2C$, respectively. Here, $C$ denotes a channel dimension, which equals 256 in the SSF-Net. The subsequent stages, Stage 3 and Stage 4, repeat the process without downsampling and produce output with the resolutions of $\frac{H}{4} \times \frac{W}{4} \times 4C$ and $\frac{H}{4} \times \frac{H}{4} \times 8C$, respectively. Each $S^2$CB within the stages consists of two $S^2$C layers, along with batch normalization (BN) and ReLU activation layers. As shown in Figure \ref{fig:3}, the $S^2$C layer utilizes 2-D depth-wise separable convolution to extract spatial features and 3-D convolution to extract multi-spectrum features. The sliding 3-D convolutional kernels establish contextual relationships between multiple continuous spectral bands and capture spectral signatures of substances. This allows the $S^2$CB to generate a joint spatial-spectral hierarchical representation. Additionally, the use of skip connections in the $S^2$CB helps mitigate issues such as vanishing gradients and facilitates information flow across different layers. In the RGB branch of our network, we utilize ResNet-50 as the feature extraction backbone. This backbone is pre-trained on RGB tracking datasets, specifically those used in the Siamese tracking framework proposed by Chen \textit{et al.} \cite{chen2020siamese}. By leveraging the pre-trained knowledge in the RGB domain, we can transfer this knowledge to the task of HS and RGB fusion tracking. The benefit from the expertise and insights gained from previous research in RGB tracking can enhance the performance and effectiveness of the proposed network.

\subsection{Spectral Attention Fusion Module}

Fusion tracking has been proven to be an effective method for leveraging the unique characteristics of both HS and RGB data. By combining the detailed visual texture information from RGB imagery with the rich spectral information from HS data, fusion trackers can achieve superior performance. However, existing methods often fail to adequately adjust spectral information during feature fusion, and they also lack comprehensive modeling of dependency relationships between spectral channels. Therefore, an SAFM is proposed to enable more refined adjustments for spectral information during feature fusion. By incorporating a channel attention mechanism, the SAFM allows for a weighted fusion of features from the two modalities, which ensures that the spectral information is appropriately considered.

\begin{figure}
    \centering
    \includegraphics[width=7cm,height=6.5cm]{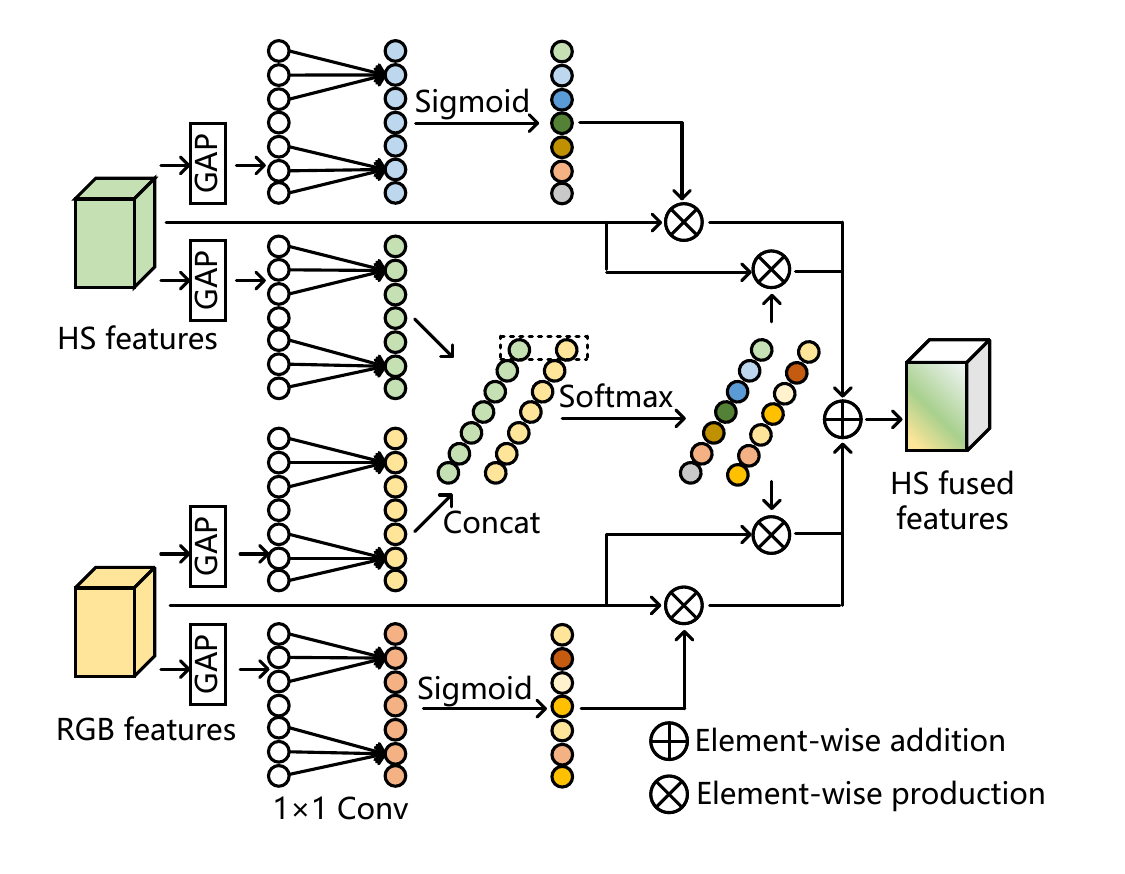}
    \caption{The architecture of the SAFM. It consists of a global average pooling layer to generate the feature vectors, and several $1 \times 1$ convolution layers to obtain the cross-modality and intra-modality attention weights.}
    \label{fig:5}
\end{figure}

In recent HS object tracking methods, feature fusion has been achieved through a spectral attention mechanism. However, most of the feature refinement is carried out within modalities, without fully considering the cross-modality interaction relationships and the complementary nature of modality features. To address these limitations, SAFM is designed to model the feature relationships within and between modalities. As shown in Figure \ref{fig:5}, a global average pooling is first used to aggregate the spatial information into a channel-wise global feature, which is useful for capturing the overall presence of features in the input features. Let $R_f \in {\mathbb{R}^{H \times W \times C}}$ and $H_f \in {{\mathbb{R}}^{H \times W \times C}}$ be the extracted features from RGB and HS images, respectively and the global average pooling is utilized to generate features $R_a \in {\mathbb{R}^{1 \times C}}$ and $H_a \in {\mathbb{R}^{1 \times C}}$. Then, intra-modality spectral embedding and inter-modality spectral embedding are achieved through 1-D convolution with different parameters. To capture local cross-channel interaction information and account for the similarity of adjacent spectra, the size of the 1-D convolutional kernel is fixed. This allows effective modeling of the relationships between channels. It is formulated by
\begin{equation}
\begin{gathered}
R_e^{intra} = C1D_{1}({R_a}), \hfill \\
H_e^{intra} = C1D_{1}({H_a}), \hfill \\
R_e^{inter} = C1D_{2}({R_a}), \hfill \\
H_e^{inter} = C1D_{2}({H_a}), \hfill \\
\end{gathered}
\end{equation}
where $C1D_{1}$ and $C1D_{2}$ are two different $1 \times 1$ convolution layers. For the intra-modality attention, with the spectral embedding features, the spectral attention weights are obtained through the $Sigmoid$ activation function, which is formulated by 
\begin{equation}
\begin{gathered}
{R_W^{intra}} = Sigmoid(R_e^{intra}), \hfill \\
{H_W^{intra}} = Sigmoid(H_e^{intra}). \hfill \\
\end{gathered}
\end{equation}

The intra-modality refined features can be obtained from the product of input features and attention weights, which are expressed as:
\begin{equation}
\begin{gathered}
R_{refined}^{intra} = {R_W^{intra}} \otimes {R_f}, \hfill \\
H_{refined}^{intra} = {H_W^{intra}} \otimes {H_f}, \hfill \\
\end{gathered}
\end{equation}
where $\otimes$ denotes the element-wise product. For the inter-modality attention, the RGB embedding $R_e^{inter} \in {{\mathbb{R}}^{1 \times C}}$ and HS embedding $H_e^{inter} \in {{\mathbb{R}}^{1 \times C}}$ are concatenated to form feature $E \in {{\mathbb{R}}^{2 \times C}}$, and the $Softmax$ activation function is used to generate the attention weight $W \in {{\mathbb{R}}^{2 \times C}}$ of the concatenated dimension, which represents dynamic guidance of the network on multi-modality correlation relationships. It is formulated by:
\begin{equation}
W = Softmax(E).
\end{equation}

Then, the inter-modality refined features between the RGB and HS modalities are obtained by the following formula:
\begin{equation}
\begin{gathered}
  R_{refined}^{inter} = W[0] \otimes {R_f}, \hfill \\
  H_{refined}^{inter} = W[1] \otimes {H_f}, \hfill \\
\end{gathered}
\end{equation}
where $W[0] \in {{\mathbb{R}}^{1 \times C}}$ and $W[1] \in {{\mathbb{R}}^{1 \times C}}$ denote the first and the second row vectors of $W$, respectively. Finally, the intra-modality and inter-modality features are added together to obtain the HS fused feature representation. As shown in Figure \ref{fig:2}, the SAFM focuses on the last three stages of the RGB and HS backbone networks and performs feature fusion between Stages 2, 3, and 4. Finally, the HS fused feature is input to the SAAM for predicting the object motions.

\subsection{Spectral Angle Awareness Module}

The existing methods only use the classification and localization branches in the existing RGB domain and calculate the depth-wise similarity of each feature \cite{li2019siamrpn++} to predict the position and size of the object. To fully utilize the spectral information in HS, a spectral angle awareness module is proposed.

1) Spectral angle mapping revisit. The spectral angle mapping (SAM), initially introduced by \cite{kruse1993spectral}, is a widely used measure in remote sensing \cite{chang2021hyperspectral, zeng2022sparse} to assess the similarity between spectral feature vectors. In the HS object detection field, SAM calculates the similarity between an object pixel and a test pixel by first calculating their corresponding mapping matrix. This matrix is then used to compute the cosine of the angle between the spectral vectors of the pixels. The cosine value serves as an effective measure of similarity, with a higher cosine value indicating a closer match to the spectral signature of the objects being compared. The specific details are as follows.

Assuming that the spectral feature $\hat S \in {{\mathbb{R}}^{H \times W \times C}}$ is a three-dimensional data cube, and it is first transformed into a two-dimensional matrix $S \in {{\mathbb{R}}^{HW \times C}}$. The original mapping matrix $M$ is obtained by:
\begin{equation}
{M} = \frac{1}{{{H}{W}}}({S^T}S),
\end{equation}
which is the estimate of the second-order statistical covariance matrix of $S$. Then, the mapping matrix ${M}^{ - 1/2}$ is obtained by:
\begin{equation}
{M}^{ - 1/2} = Q{V^{ - 1/2}}Q^T,
\end{equation}
where $Q$ is a unitary matrix, and $V$ is the eigenvalue diagonal matrix of $M$. With the mapping matrix ${M}^{ - 1/2}$, the spectrum $t$ of an object and a sparse tensor $S$ are projected into the same subspace. The cosine similarity between the spectrum of the object $t$ and feature map $S$ is formulated as:
\begin{equation}
\cos _{(i,j)}^ \wedge  = \frac{{\left\langle {M^{ - 1/2}{S_{(i,j,:)}},M^{ - 1/2}t} \right\rangle }}{{||M^{ - 1/2}{S_{(i,j,:)}}|| \cdot ||M^{ - 1/2}t||}},
\end{equation}
where $(i,j)$ represents the pixel location of feature maps $S$ to be detected. With the cosine similarity, the influence of background clutter can be mitigated, and the performance of object detection can be improved.

\begin{figure}
    \centering
    \includegraphics[width=9.0cm,height=8.5cm]{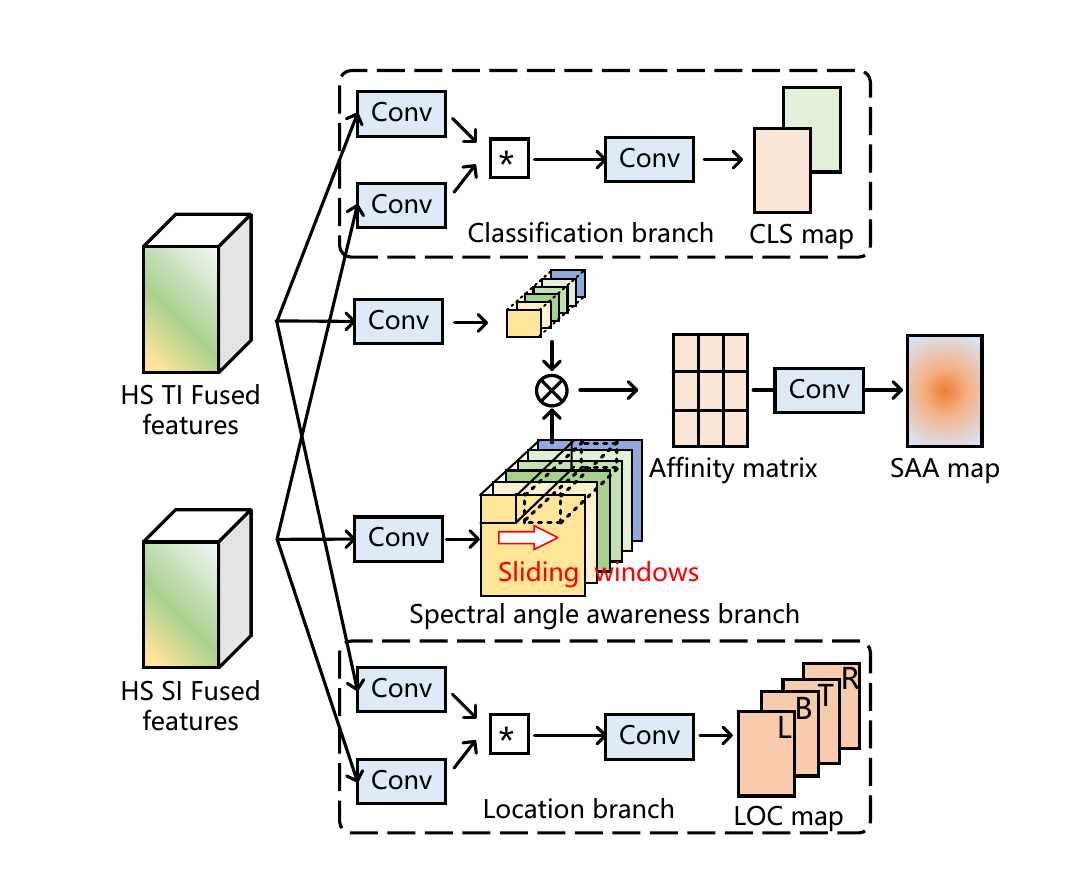}
    \caption{The architecture of the SAAM includes the CLS branch, LOC branch, and SAA branch. The CLS and SAA branches predict the position of the objects together, and the LOC branch predicts the size of the objects.}
    \label{fig:6}
\end{figure}

2) Spectral angle awareness branch. The structure of SAAM is shown in Figure \ref{fig:6}. To improve upon the existing formulation, we have reformulated Equation (8) and presented a simplified formula for computing similarity in the SAM. This simplified equation allows for a more efficient and effective calculation, which is given by:
\begin{equation}
\cos {\theta _{(i,j)}} = \frac{{\left\langle {f{{(z)}_{(i,j,:)}},f{{(x)}_{(i,j,:)}}} \right\rangle }}{{||f{{(z)}_{(i,j,:)}}|| \cdot ||f{{(x)}_{(i,j,:)}}||}},
\end{equation}
where $z$ and $x$ represent TI and SI of the HS modality, respectively, $f(z)$ and $f(x)$ are the feature maps extracted by the same $f$, which is a series of feature extraction and regularization functions. In this work, $f$ is the feature extraction network $S^2$FB. In the field of HS tracking, both the classification and localization heads rely on region-level similarity assessments to make their predictions. The features of the TI and SI extracted by the tracker are denoted as $f(z)$ and $f(x)$, respectively. To perform similarity assessments, the feature map $f(x)$ undergoes depth-wise separable convolutions using $f(z)$ as the convolution kernel. This operation is given by:
\begin{equation}
Sim = f(z) * f(x),
\end{equation}
where $*$ represents depth-wise cross-correlation layer \cite{li2019siamrpn++}. It can be seen as an extended region-level version of the SAM, all of which aim to obtain similarity through vector operations. The idea of spectral angle can seamlessly integrate into the tracking prediction head. To achieve this integration, an embedding layer is employed to project the feature maps $f(z)$ and $f(x)$ onto a subspace. This subspace projection enables the calculation of cosine similarity between the two feature maps, expressed as:
\begin{equation}
\begin{gathered}
  f_z^{embed} = {E_z}(f(z)), \hfill \\
  f_x^{embed} = {E_x}(f(x)), \hfill \\
\end{gathered}
\end{equation}
where $E_z$ and $E_x$ denote the embedding layers used by $f(z)$ and $f(x)$, respectively. The embedded feature $f_z^{embed}$ is utilized as a sliding convolution kernel to convolve with $f_x^{embed}$, resulting in an output map. Consequently, this procedure facilitates the calculation of the region-level affinity matrix between $f_z^{embed}$ and $f_x^{embed}$. Subsequently, a predicted map of spectral angular affinity $M_{SAA}$ is derived using a $1 \times 1$ convolutional layer. The detailed formula is presented as:
\begin{equation}
M_{SAA} = CV2D_{1\times1}(Sim(f_z^{embed}, f_x^{embed})),
\end{equation}
where $Sim$ denotes the 2-D convolutional calculation, and $CV2D_{1\times1}$ denotes the $1 \times 1$ convolutional layer.

\begin{figure}
    \centering
    \includegraphics[width=6.5cm,height=3.5cm]{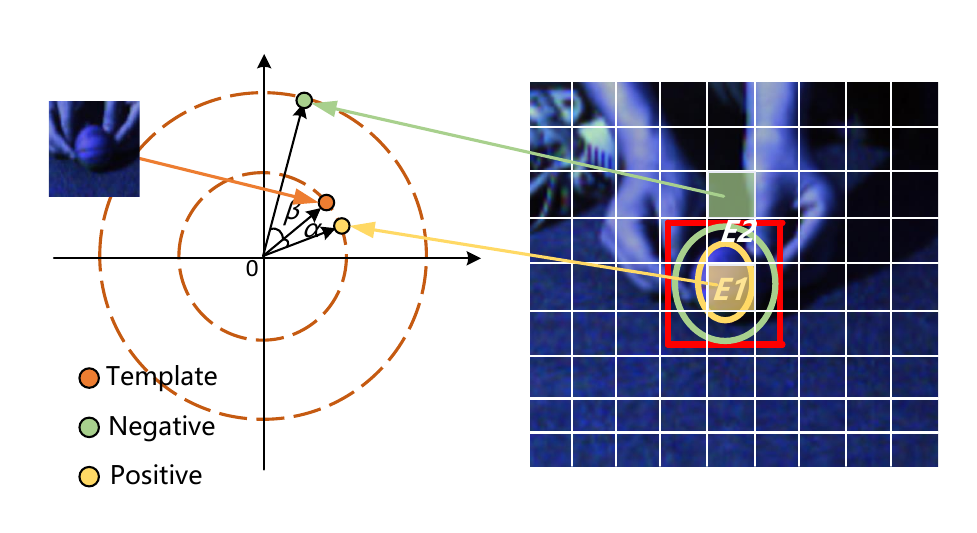}
    \caption{The architecture of the SAAL. During optimization, the similarity of positive sample pairs learned by the network increases, while the similarity of negative sample pairs decreases.}
    \label{fig:11}
\end{figure}

To train the SAAM more effectively, the SAAL is proposed. The positive and negative sample regions, as \cite{chen2020siamese}, are constructed to transfer the triplet optimization \cite{hermans2017defense} to the HS tracking. The core idea behind triplet loss is to take three data samples at a time, typically referred to as an “anchor”, a “positive”, and a “negative”. The anchor and positive samples belong to the same class or depict the same object, while the negative samples come from a different class. Therefore, the SAAL is designed to maximize the similarity between the anchor and the positive while minimizing the similarity between the anchor and the negative. As shown in Figure \ref{fig:11}, the ground-truth bounding box of SI is marked with the red box, and two ellipses $E1$ and $E2$ are constructed. The position $(i,j)$ inside $E1$ is marked as positive $p$, and the position $(i,j)$ outside $E2$ is marked as negative $n$. We use the features of TI extracted by the tracker as anchors to form positive pairs with positive regions in SI and negative pairs with negative regions in SI. The SAAL is formulated by:
\begin{equation}
{L_{SAAL}} = \sum_{\substack{p \in E1\\n \in \complement (E2)}} {[Sim(a,p) - Sim(a,n)]},
\end{equation}
where $p \in E1$ denotes the point inside $E1$ of the $M_{SAA}$, and ${n \in \complement (E2)}$ is the complementary point of $E2$, i.e., outside $E2$. $a$ denotes the template image, and $Sim(a,p)$ and $Sim(a,n)$ are the cosine similarities of positive pairs and negative pairs, respectively. During the optimization process, the network learns to increase the similarity between positive sample pairs while decreasing the similarity between negative sample pairs. This means that the angle between the feature vectors of positive sample pairs in the feature space becomes smaller, while the angle between the feature vectors of negative sample pairs becomes larger. This intuitive understanding is illustrated in Figure \ref{fig:1}(c). During inference, SAAM assists the trackers in enhancing the response of objects within TI as it appears in SI. This enhancement facilitates the precise localization of the object within the region of similarity shared between TI and SI.

3) Classification and location branch. The classification and location heads are widely used to predict the location and size of objects \cite{chen2020siamese}. The former outputs the CLS map ${M_{CLS}} \in {\mathbb{R}^{H \times W \times 2}}$ to estimate whether the pixels belong to targets or background, and the latter outputs the four-channel LOC map ${M_{LOC}} \in {\mathbb{R}^{H \times W \times 4}}$ to calculate the offset from the predicted location to four sides to ground-truth box. To train the SSF-Net effectively, we employ a combination of the SAAL, cross-entropy loss, and the intersection over union (IOU) loss \cite{yu2016unitbox}. This joint training approach allows the network to optimize both the classification and localization performance. The SAAL is incorporated to enhance the learning of spectral angle awareness, while the cross-entropy loss and the IOU loss contribute to the overall training objective. It is formulated as:
\begin{equation}
L = \alpha {L_{CLS}} + \beta {L_{LOC}} + \gamma {L_{SAAL}},
\end{equation}
where $\alpha$, $\beta$, and $\gamma$ denote the trade-off parameters. Here, we set $\alpha=1$, $\beta=2$ following \cite{liu2022siamhyper}, and $\gamma=1$ is set to 1 for all experiments. Then, to ensemble the prediction from HS and RGB modalities, a weighted prediction method is adopted to learn the weights of the two modalities in a self-adaptive way. The weighted fusion prediction $M_{CLS}$ and $M_{LOC}$ are obtained from:
\begin{equation}
\begin{gathered}
{M_{CLS}} = {\lambda _1}M_{CLS}^{HS} + {\lambda _2}M_{CLS}^{RGB}, \hfill \\
{M_{LOC}} = {\lambda _1}M_{LOC}^{HS} + {\lambda _2}M_{LOC}^{RGB}, \hfill \\
\end{gathered}
\end{equation}
where ${\lambda _1}$ and ${\lambda _2}$ are the trainable parameters. With the weighted prediction, the CLS and LOC maps integrate the information from two branches to enhance the accuracy of object localization. Finally, considering the widespread utilization of ensemble prediction in existing methods, we aggregate the prediction maps from the final three stages of both the RGB and HS backbone networks. Furthermore, through the joint calculation of the loss function, the proposed method can simultaneously predict the class and location of objects during the inference stage. This integrated approach enhances the efficiency and effectiveness of object prediction.

\section{Experiments}

\subsection{Experiment Settings}

\textit{Datasets.} The HOTC-2020 dataset \cite{xiong2020material} is used to validate the effectiveness of the proposed method. The dataset is divided into a training set comprising 40 videos and a testing set consisting of 35 videos. To capture different objects, both RGB and HS cameras are employed for simultaneous recording, resulting in each video containing three types of modality data: RGB, HS, and generated false-color images. The HS image cubes within the dataset consist of 16 spectral bands. Furthermore, the dataset contains 11 challenging factors, including background clutter (BC), deformation (DEF), fast motion (FM), illumination variation (IV), in-plane rotation (IPR), low resolution (LR), motion blur (MB), out-of-occlusion (OCC), plane rotation (OPR), out-of-view (OV), and scale variation (SV). 

The HOTC-2024 is developed for the 2024 Hyperspectral Object Tracking Challenge and features three distinct band configurations: 15 bands (RedNIR), 16 bands (VIS), and 25 bands (NIR). Each video frame is stored as a 2D image and can be converted into a 3D image cube using publicly available algorithms. The dataset includes 218 hyperspectral videos for training and 118 for validation, with each frame annotated with a bounding box for the tracked object. We conduct experiments using datasets of 15 bands, labeled as HOTC-2024-RedNIR.

The BihoT \cite{wang2024bihot} is the first dataset used for hyperspectral camouflage object tracking, with 41,912 images captured by an IMEC XIMEA MQ022HG-IM-SM5X5-NIR camera in the 25 bands, including 22,992 images in the training set and 18,920 images in the testing set. The dataset is annotated with 9 attributes, including Background Clutter (BC), Fast Motion (FM), In-Plane Rotation (IPR), Illumination Variation (IV), Low Resolution (LR), Occlusion (OCC), Out-of-Plane Rotation (OPR), Spectral Consistency (SC), and Spectral Variation (SV).

\textit{Evaluation Metrics.} The area under the curve (AUC) of the success plot and the precision rate at the threshold of 20 pixels, which is denoted as DP\_20, are used as the evaluation metrics. The success plots are utilized to assess the Intersection over Union (IOU) between the predicted bounding box and the ground truth. These plots illustrate the proportion of frames exceeding various IOU thresholds by depicting a success curve, with the IOU threshold spanning from 0 to 1. A larger area under the curve (AUC) represents the superior tracking performance. Regarding precision metrics, they gauge the discrepancy in pixels between the centers of the predicted and ground truth bounding boxes. The precision rate curve is then constructed by plotting the proportion of frames with a center offset below a given threshold, which extends from 0 to 50 pixels. A lower center offset corresponds to enhanced tracking accuracy.

\textit{Implementation Details.} All experiments are conducted using an NVIDIA RTX 4090 GPU. For the RGB branches, we utilize a pre-trained ResNet-50 network sourced from training on the COCO \cite{lin2014microsoft}, ImageNet VID \cite{russakovsky2015imagenet}, and YouTube-BB \cite{real2017youtube} dataset as the backbone network. The SSF-Net is not initialized with any pre-trained parameters. During the training phase, the parameters of the RGB stream are fixed, while only the parameters of the HS stream are subject to training. We set the batch size to 30, and the learning rate is initialized as 0.005, applying a warm-up learning rate adjustment strategy. The weight decay and momentum are configured at 0.0001 and 0.9, respectively. 


\begin{table*}[]
\caption{Overall performance comparison of HS and visual tracker. The HS trackers deal with the HSIs, and the visual tracker runs on the false-color images. We use the red and blue fonts to highlight the top two values.}

\label{tab:5}
\resizebox{\textwidth}{!}{%
\begin{tabular}{c|ccccc|cccc}
\hline
                            & \multicolumn{5}{c|}{HS tracker}                                                                                                           & \multicolumn{4}{c}{Visual tracker}                                                                                    \\ \hline
\multicolumn{1}{l|}{Method} & \multicolumn{1}{l}{SSF-Net} & \multicolumn{1}{l}{BAE-Net\cite{li2020bae}} & \multicolumn{1}{l}{SEE-Net\cite{li2023learning}} & \multicolumn{1}{l}{SPIRIT\cite{chen2023spirit}} & \multicolumn{1}{l|}{SST-Net\cite{li2021spectral}} & \multicolumn{1}{l}{TransT\cite{chen2021transformer}} & \multicolumn{1}{l}{SiamGAT\cite{guo2021graph}} & \multicolumn{1}{l}{SimTrack\cite{chen2022backbone}} & \multicolumn{1}{l}{OSTrack\cite{ye2022joint}} \\ \hline
AUC                         & \textcolor{red}{0.680}                       & 0.606                       & 0.654                       & \textcolor{blue}{0.679}                       & 0.623                        & 0.633                      & 0.581                       & 0.599                        & 0.557                       \\
DP\_20                      & \textcolor{red}{0.939}                       & 0.878                       & 0.907                       &  \textcolor{blue}{0.925}                       &0.917                        & 0.879                      & 0.827                       & 0.845                        & 0.815                       \\ \hline
\end{tabular}%
}
\end{table*}

\begin{table*}[]
\caption{Overall performance comparison of HS trackers on the HOTC-2024-RedNIR and BihoT datasets. We use the red and blue fonts to highlight the top two values.}
\label{tab:hotc24}
\centering
\renewcommand{\arraystretch}{1.5}
\scalebox{1.0}{
\begin{tabular}{cccccccc}
\hline
Datasets & Metrics & SEE-Net\cite{li2023learning}   & SiamHYPER\cite{liu2022siamhyper} & SENSE\cite{chen2024sense} & MMF-Net\cite{li2024material} & SSF-Net\\ \hline
 \multirow{4}{*}{HOTC-2024-RedNIR}           & AUC        & 0.332 & 0.328   & \textcolor{blue}{0.351}   & 0.324   & \textcolor{red}{0.357}           \\ \cline{2-7}
 & DP\_20      & 0.433 & 0.428   & \textcolor{red}{0.469}   & 0.424   & \textcolor{blue}{0.463} \\ \cline{2-7}
  & FLOPs        & 297.99G & \textcolor{blue}{121.70G}   & 287.74G   & 656.67G   & \textcolor{red}{83.37G}           \\ \cline{2-7} 
   & FPS        & 12.06 & \textcolor{blue}{12.62}   & 11.55   & 3.75   & \textcolor{red}{13.05}               \\ \hline
                            
\multirow{2}{*}{BihoT}           & AUC                 & 0.315 & 0.403   & 0.454   & \textcolor{blue}{0.475}   & \textcolor{red}{0.497}            \\ \cline{2-7} 
             & DP\_20              & 0.422 & 0.539   & 0.576   & \textcolor{blue}{0.580}   & \textcolor{red}{0.668}

             \\ \hline
 \end{tabular}%
 }
 \end{table*}

\begin{figure} [!htb] 
\centering
\includegraphics[width=8cm,height=8cm]{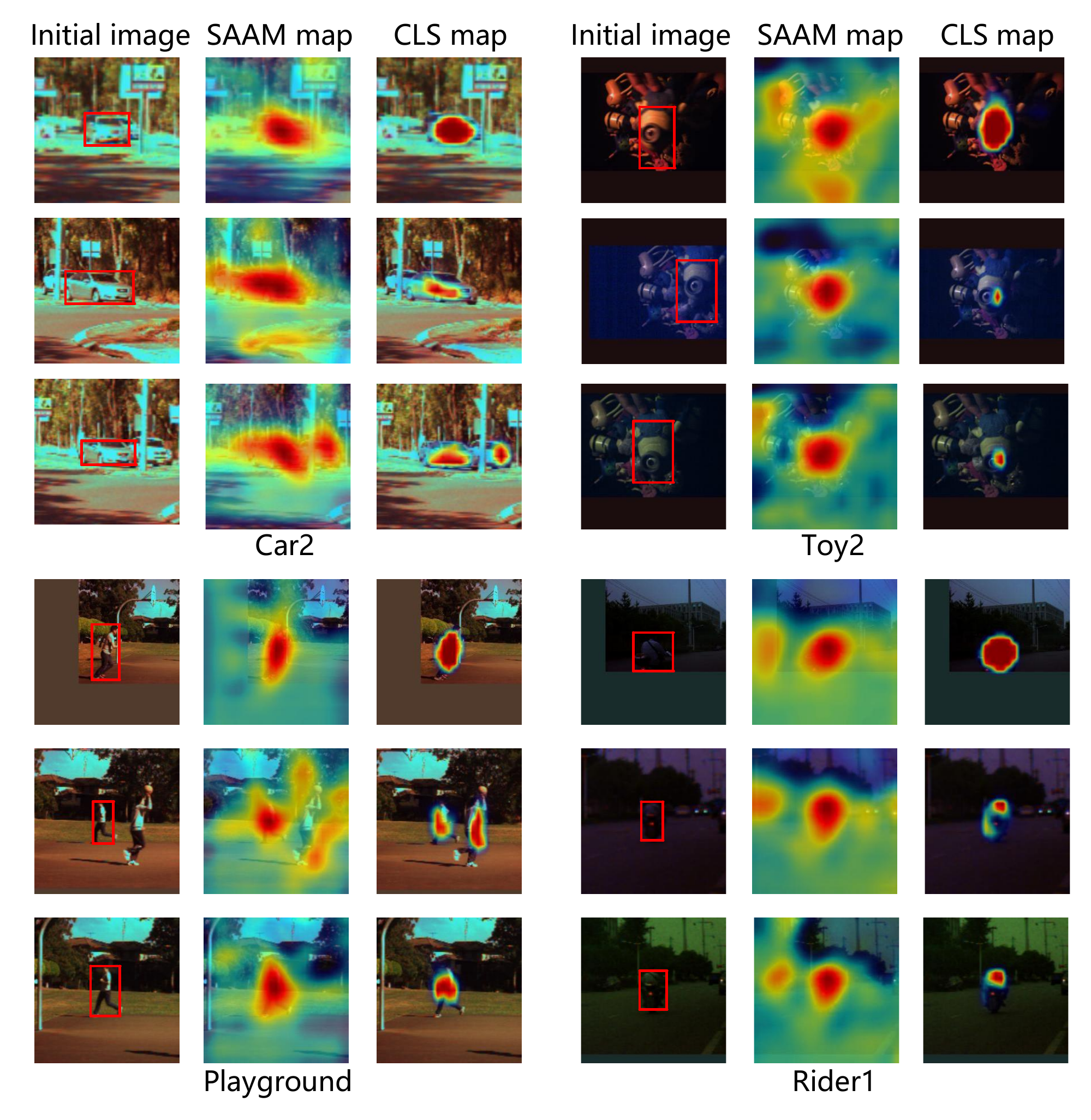}
\caption{The visualization of heatmaps on the HOTC-2020 dataset. The red color represents high response while the blue color represents low response.}
\label{fig:9}
\end{figure}

\subsection{Comparison with State-of-the-arts}

\textit{Comparison with HS Trackers.} We first compare the proposed SSF-Net with advanced HS trackers, including band regrouping-based methods SEE-Net, BAE-Net, SST-Net, and band aware-based method SPIRIT. BAE-net initially learned spectral weights in the form of channel attention, and it regrouped the HS cubes to multiple three-channel false-color images based on the spectral weight, thereby transferring existing RGB trackers to process HS tracking. SEE-Net learned the weights between bands in the form of an autoencoder. SST-Net built a temporal-aware false-color image generation framework. SPIRIT utilized both channel attention and an autoencoder-like structure for band awareness while also considering both intra-band and inter-band interactions. It is worth noting that we reproduced SEE-Net, and the results of SPIRIT are not available, thus we obtain the accuracy from the corresponding paper directly without attribute-based and visualized tracking results.

The experimental results are listed in Table \ref{tab:5}. Without consideration of HS feature extraction, the performance of BAE-Net using only spectral attention is 0.606 of the AUC score and 0.878 of DP\_20. The tracking performance of SST-net, a band regrouping method that considers spatial, spectral, and temporal information simultaneously, can achieve 0.623 of AUC score and 0.917 of DP\_20, which compensates for the lack of spectral attention. SEE-Net regroups bands from the perspective of an autoencoder, further achieving 0.654 of the AUC score and 0.907 of DP\_20. It can be seen that using only the channel attention mechanism or autoencoder-like structures has limited ability in mining band relationships, which yields the general tracking performance. Band grouping-based methods are not powerful in extracting robust HS features, and using existing RGB trackers to infer the modeling of the spectral relationships limits the performance of HS tracking. By jointly considering the spectral attention mechanism and autoencoder structure, SPIRIT provides more effective tracking results with an AUC score of 0.679 and DP\_20 of 0.925. The perception of spectral relationships and the Transformer-based prediction head further improve the performance. Our method not only considers the fusion of spectral and spatial information but also designs a spectral sensing module for tracking and prediction, and the experimental results achieve the best, with the AUC score of 0.680 and DP\_20 of 0.939. This shows the effectiveness of SSF-Net, where $S^2$FB and SAFM can extract more discriminative features, while SAAM can more accurately predict the position of objects.

Besides, to prove the effectiveness of our methods further, we conduct comparative experiments on the HOTC-2024-RedNIR and BihoT datasets. The experimental results are listed in Table \ref{tab:hotc24}. Our method outperforms most existing methods, achieving an AUC of 0.357 and a DP\_20 of 0.463 on the HOTC-2024-RedNIR dataset, and our method surpasses other methods in terms of AUC and DP\_20 on the BihoT dataset. This improvement is attributed to our proposed modules ($S^2$FB, SAFM, and SAAM), which enable the model to form joint spatial-spectral features of the objects. Additionally, for camouflaged objects with greater tracking difficulty, our method leverages SAAM to perceive spectral similar regions between the template and the search features, further enhancing tracking performance, which demonstrates the effectiveness and generalizability of our proposed methods.

\begin{figure} [!htb] 
\centering
\includegraphics[width=8cm,height=8cm]{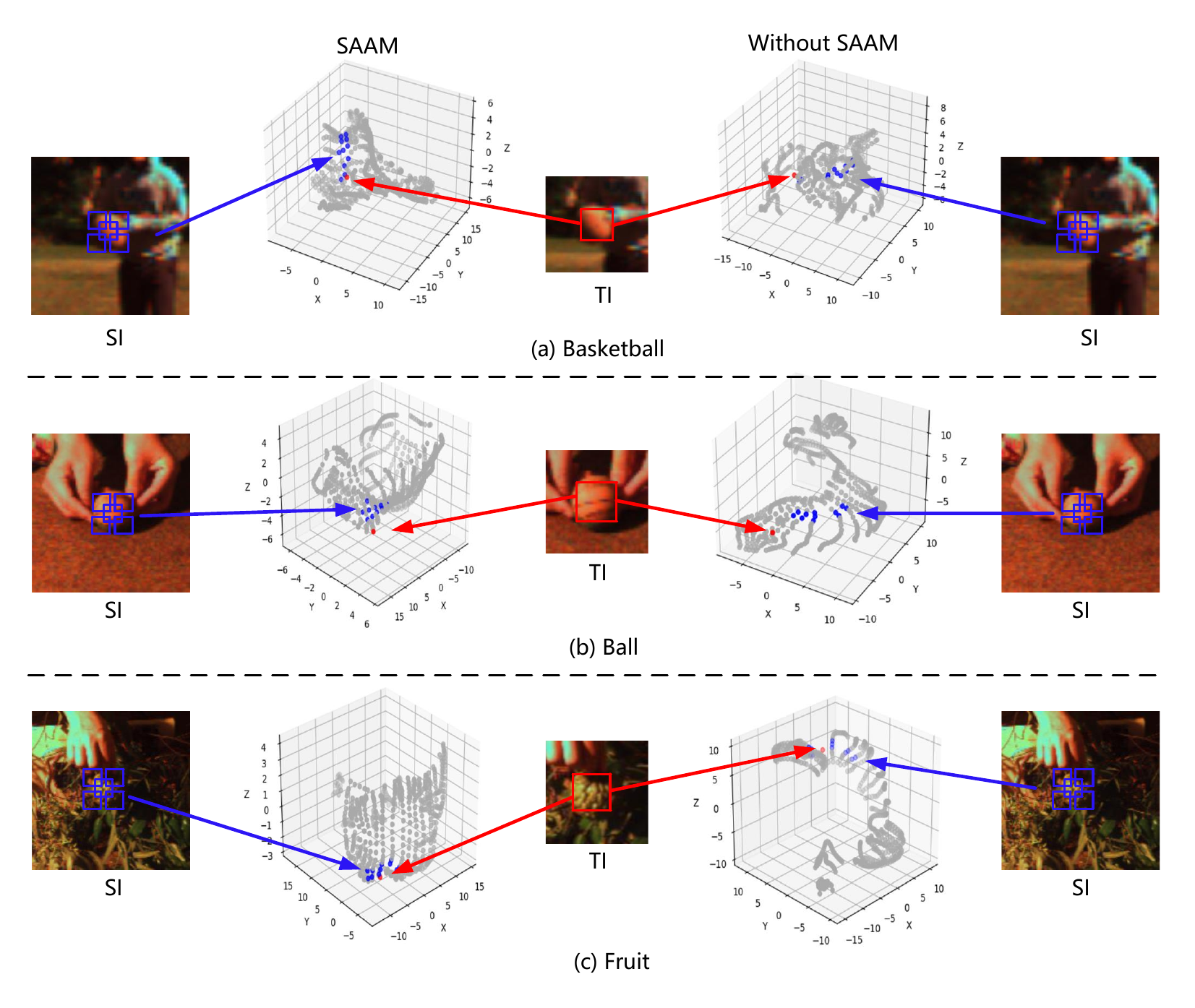}
\caption{The feature distribution visualization of SAAL. The red points denote the TI samples, and the blue and gray points denote the positive and negative SI samples.}
\label{fig:10}
\end{figure}

\textit{Comparison with RGB Trackers.} In addition, we also compare the SSF-Net with advanced RGB trackers, including SiamGAT, OSTrack, SimTrack, and TransT. The false-color images of the HOTC-2020 dataset are used to perform experiments. As shown in Table \ref{tab:5}, the SSF-Net provides better performance compared with the RGB tracker. Due to the difficulty in handling factors such as lighting changes and background interference, the RGB trackers are generally weaker than HS trackers in processing challenging scenes.

\textit{Visualization Analysis.} To better demonstrate the performance of our tracker and the effectiveness of the proposed SAAM and SAAL, we conduct several visualization experiments in this section.

\textit{1) Heatmap Visualization.} To demonstrate the effectiveness of our proposed SAAM intuitively, we utilize the  Grad-CAM++ \cite{chen2018gradnorm} for feature visualization of classification heads and SAAM. The brighter the color, the higher the level of attention received by the network, which reflects the contribution of tracking performance in this region. The four scenarios are selected including \textit{car2}, \textit{toy2}, \textit{playground} and \textit{rider1}. The experimental results are shown in Figure \ref{fig:9}. The first and fourth columns are the original images. The ground-truth bounding box is framed by red lines. The second and fifth columns are SAA maps, and the third and sixth columns are CLS maps. The brighter the color, the higher the attention received by the network, reflecting the contribution of tracking performance in this region. Due to the visually similar cars and people in the \textit{car2} and \textit{playground} scenes, traditional CLS prediction heads may be confused with tracking objects, as shown in the CLS map of \textit{car2} in the third line and the CLS map of \textit{playground} in the second line. The prediction head also responds to similar objects that appear, which may be detrimental to improving tracking performance. The SAAM focuses on spectral similarity calculation, which can handle the problem of similar appearances of different objects. We can see that although SAAM also responds to other similar objects, such as the SAA map of \textit{car2} in the third row and the SAA map of \textit{playground} in the second row, the area with the highest response is still the tracking object, reflecting the robustness of the SAAM algorithm in handling similar objects. In addition, it can be seen from \textit{toy2} and \textit{rider1} scenarios that SAAM has a large receptive field and can perceive entire objects, avoiding the local overfitting of objects.

\begin{figure*} [!htb]
\centering
\includegraphics[width=16.5cm,height=9.5cm]{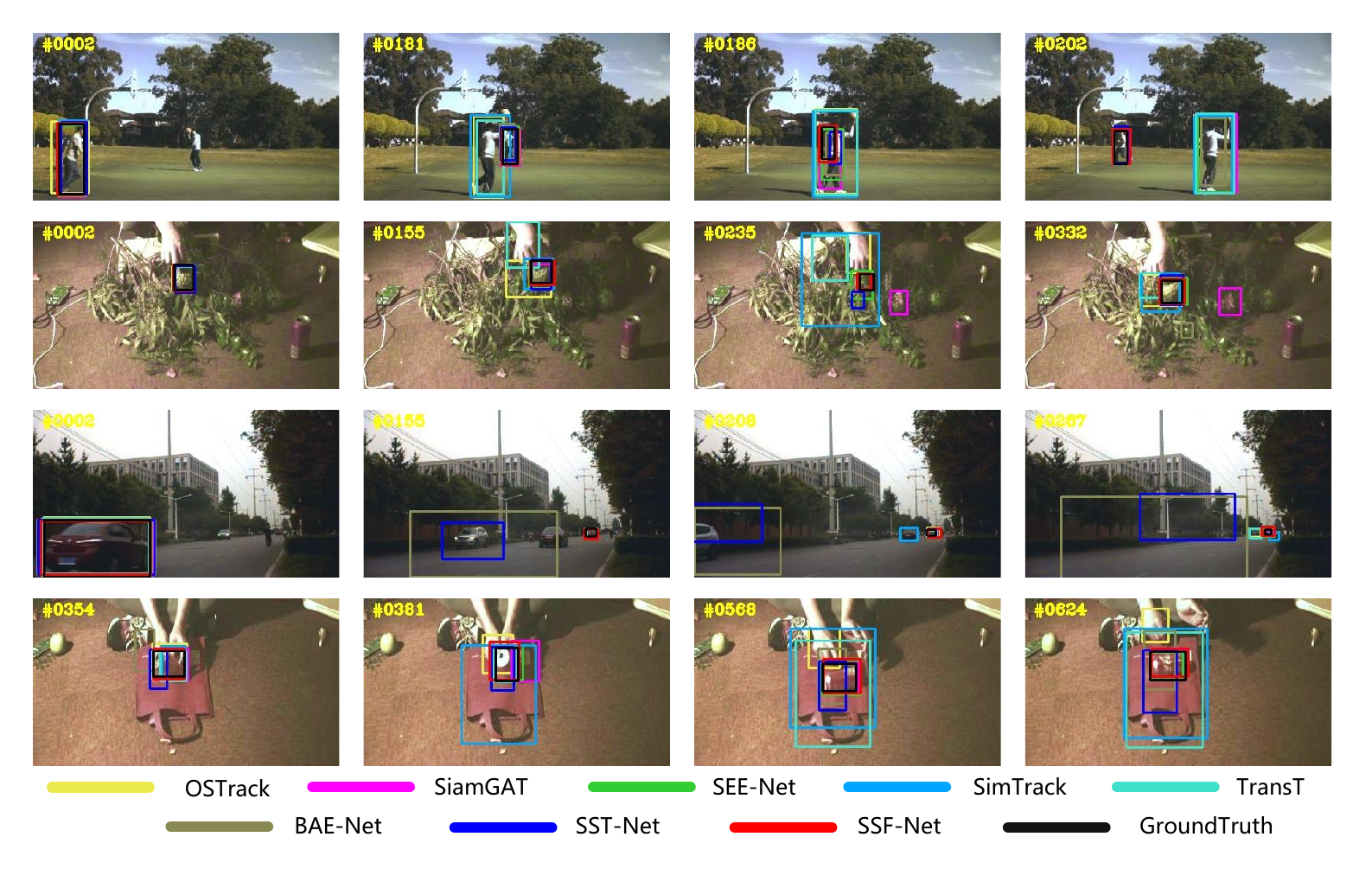}
\caption{Tracking results on the HOTC-2020 datasets of the false-color images under the \textit{playground}, \textit{forest}, \textit{car3}, and \textit{coke} scenarios.}
\label{fig:8}
\end{figure*}

\begin{figure*} [!htb] 
\centering
\includegraphics[width=16.5cm,height=8.5cm]{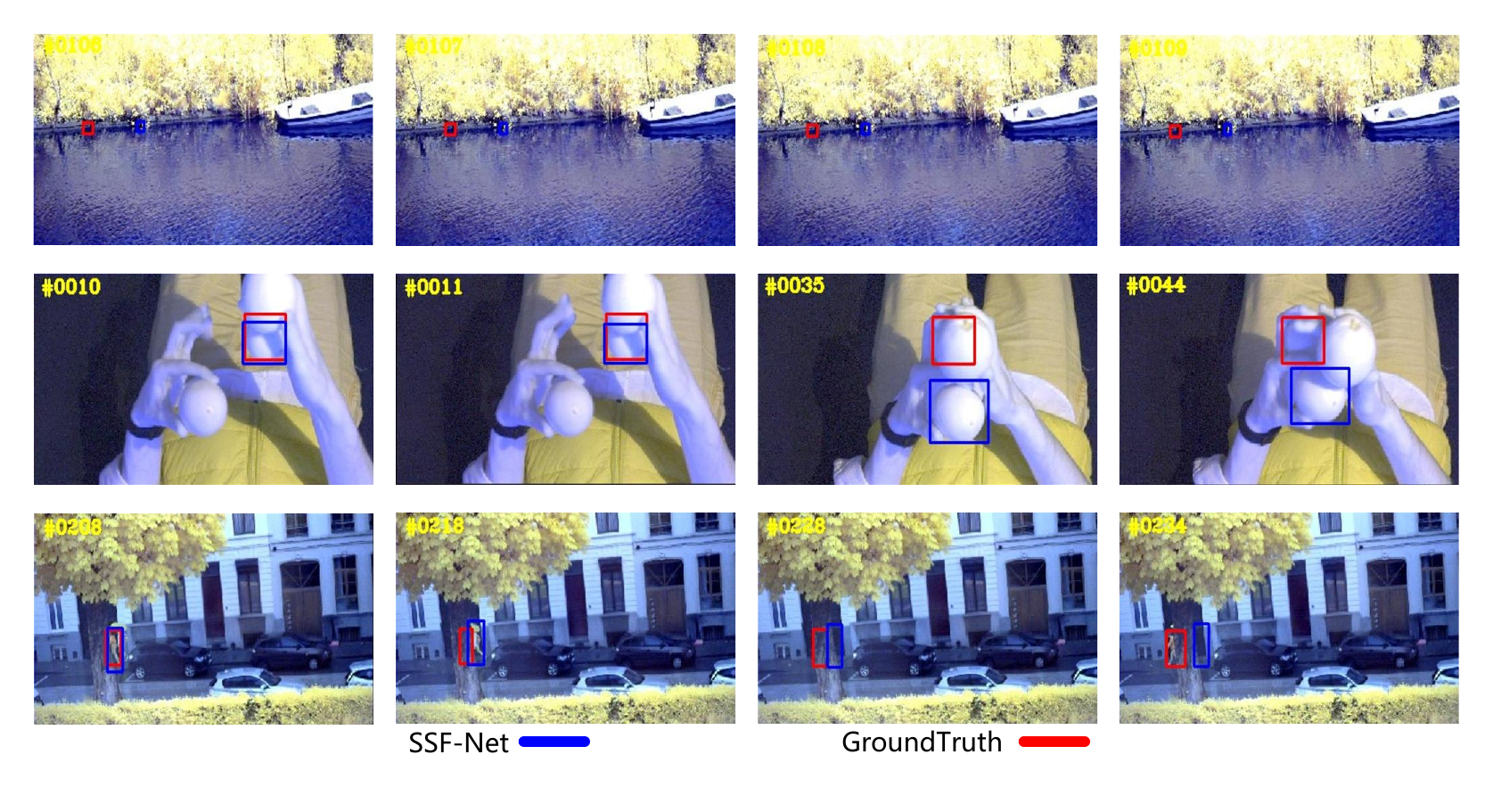}
\caption{Failure cases on the HOTC-2024-RedNIR datasets of the false-color images under the \textit{football1}, \textit{oranges5}, and \textit{rainystreet12} scenarios.}
\label{fig:failure}
\end{figure*}

\begin{figure*} [!htb] 
\centering
\includegraphics[width=17cm,height=13cm]{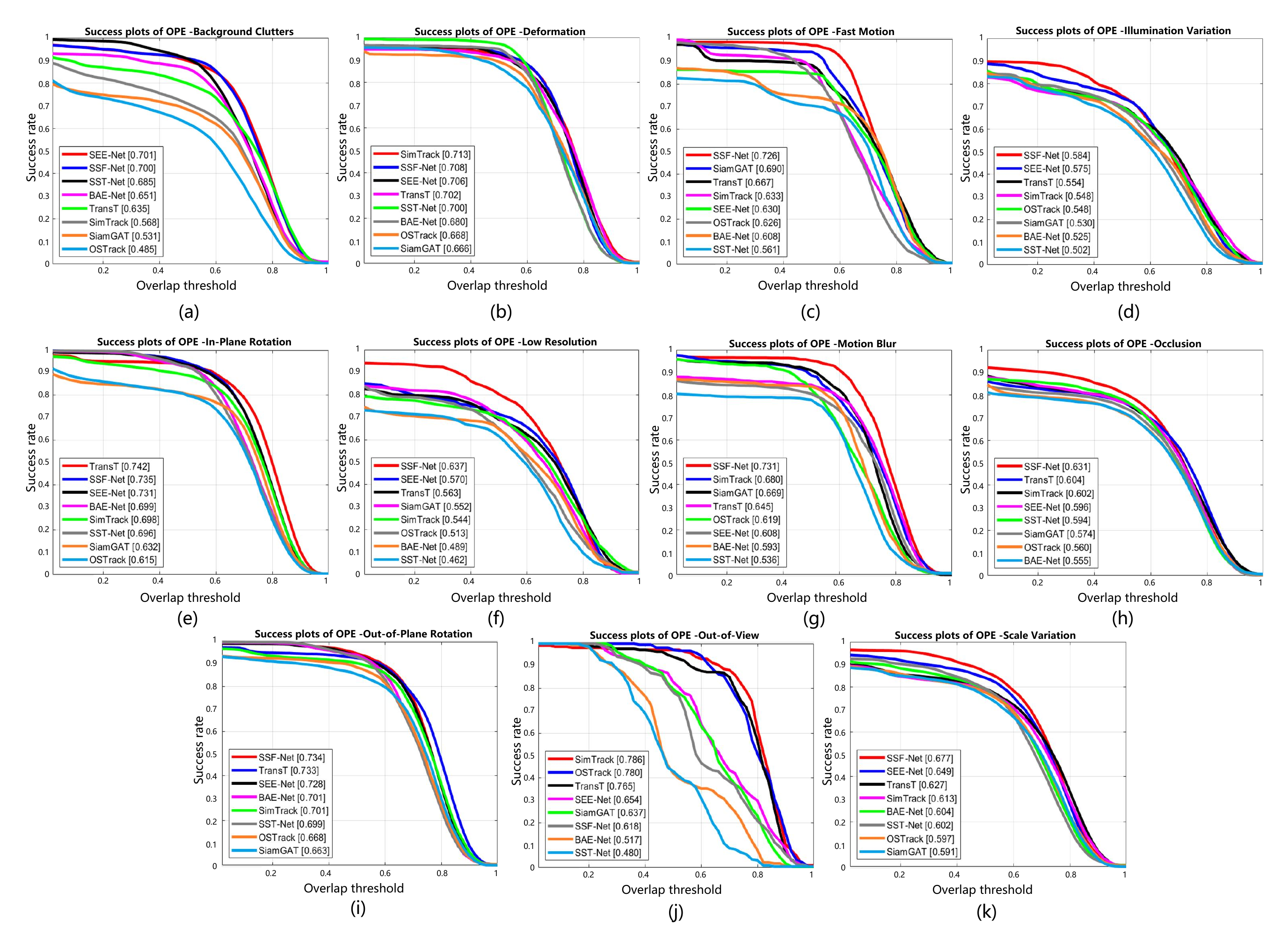}
\caption{Attribute-based comparison of success plots. (a) Success plot of background clutter (BC). (b) Success plot of deformation (DEF). (c) Success plot of fast motion (FM). (d) Success plot of illustration Variation (IV). (e) Success plot of in-plane rotation (IPR). (f) Success plot of low resolution (LR). (g) Success plot of motion blur (MB). (h) Success plot of occlusion (OCC). (i) Success plot of out-of-plane clutters (OPR). (j) Success plot of out-of-view (OV). (k) Success plot of scale variation (SV).}
\label{fig:7}
\end{figure*}

\textit{2) Feature Distribution Comparison.} To prove the validity of our proposed SAAL, we also select frames from three different clips and use the \textit{t-SNE} methods to visualize the spectral features of TI and different region features of SI. As shown in Figure \ref{fig:10}, the blue box in SI represents the area to be detected where the tracked object appears, while the red box in TI represents the object to be tracked, corresponding to blue and red points in the feature space, respectively. Local features that do not belong to the area to be detected are represented by gray points. After training with SAAL, the TI and SI samples can come closer in the feature space. With the proposed SAAL, the tracker can achieve better similarity modeling ability and correctly detect the object to be tracked.

\textit{3) Tracking Results.} We select videos from four different scenarios, including \textit{playground}, \textit{forest}, \textit{car3}, and \textit{coke}, to show the visualization results of advanced RGB and HS trackers. As shown in Figure \ref{fig:8}, the ground-truth bounding box is framed by black lines. The subsequent tracking results are obtained from the first frames based on the positions of labels, which are shown from the second to the fourth columns. Firstly, for \textit{playground} scenes where occlusion is the main challenge, both the HS tracker and RGB tracker are susceptible to visual interference from similar objects, for example, frames 181 and 188, resulting in incorrect objects. Our proposed SSF-Net, due to its focus on the similarity between bands, is robust in processing visually similar objects and can effectively solve the occlusion problem of similar objects. Then, we can also see from the scene of \textit{car3} that most trackers struggle to handle the scale changes of objects, especially for tracking small objects. SSF-Net can extract more spectral information, ensuring feature integrity in the event of visual information loss, thereby maintaining tracking performance. In addition, the results of \textit{fruit} and \textit{coke} scenes also show the effectiveness of SSF-Net in dealing with object rotation and background clutter. Besides, we also visualize the failure cases on the HOTC-2024-RedNIR dataset, which are shown in Figure \ref{fig:failure}, When an object moves rapidly or has a small size (as shown in the first row), the useful spectral and spatial features tend to be compressed, reducing the discriminability of the extracted features. This degradation negatively impacts the tracking performance of the proposed method. Additionally, when tracking multiple similar objects (as shown in the second row), the similarity in spatial and spectral information may lead to ambiguous tracking clues, making it difficult to distinguish between objects accurately. This can result in tracking failure, especially in scenarios where objects exhibit high spectral resemblance.

\textit{Attribute-based Comparison.} To demonstrate the ability to handle the challenging factors of SSF-Net, Figure \ref{fig:7} shows the success plot results of 11 attribute-based experiments. It can be observed that the SSF-Net achieved the top two results in 10 out of 11 scenarios. Among them, SSF-Net significantly outperforms other trackers in scenarios where spatial information is disrupted, such as BC, SV, FM, LR, DEF, and MB, which shows the robustness of the algorithm equipped with SAAM against visual interference. Due to the fusion of spectral and visual information, the SSF-Net can respond to objects in a transient occlusion state and has superior performance in scenarios such as OCC and IV. For the two rotating scenes of IPR and OPR, although the visual features of the object have changed, the HS tracker can extract spectral features. In particular, SSF-Net captures the object based on spectral similarity, thus yielding better performance. However, the performance of SSF-Net in OV is poor because the algorithm makes it difficult to extract robust features for objects beyond the receptive field and struggles to integrate spectral and spatial information. Nevertheless, SSF-Net still achieves an overall performance of 0.680 AUC score and 0.939 DP\_20, which demonstrates the effectiveness of the proposed methods.

\subsection{Running Cost Comparison.}

To demonstrate the efficiency of our proposed method, we conduct running cost experiments on the HOTC-2024-RedNIR dataset. The experimental results using FLOPs and frames per second (FPS) as the evaluation indicators are reported in Table \ref{tab:hotc24}. All the trackers are run on a Windows machine with an NVIDIA 4090 GPU. We can see that the band regrouping-based methods (e.g., SEE-Net and SENSE) rely on the simultaneous inference of multiple false-color images, resulting in higher computational and parameter complexities compared to our proposed method, ultimately leading to lower FPS. Additionally, MMF-Net, which integrates HSI, RGB, and material information, requires extensive preprocessing steps such as spectral unmixing, significantly slowing down its tracking speed. In contrast, SSF-Net achieves high tracking performance while maintaining efficient processing, demonstrating the effectiveness and efficiency of the proposed method.

\subsection{Ablation Experiments}

\textit{Evaluation of Each Component.} To evaluate the effectiveness of the proposed component, we conduct ablation experiments on two datasets: HOTC-2020 and HOTC-2024-RedNIR. Firstly, we reproduce the bi-stream ResNet-50 network proposed by Liu \textit {et al.} \cite{liu2022siamhyper} as the baseline, which is marked as “B”. Then, we replace the ResNet-50 network of HS stream with $S^2$FB, which is annotated as “S”, and conduct experiments under the same fusion module SSATTN \cite{liu2022siamhyper}. The experimental results are listed in Table \ref{tab:1} and Table \ref{tab:hot24}. As for HOTC-2020, compared with the baseline model, $S^2$FB improves the AUC score by 0.012 and  DP\_20 by 0.008. Then, based on “S”, we replace SSATTN with SAFM to demonstrate the superiority of SAFM in intra-modality and cross-modality spectral attention, which is denoted as “F”. The AUC score yields 0.675, and DP\_20 yields 0.934, demonstrating the effectiveness of SAFM. Furthermore, the SAAM is incorporated into “F”, which is annotated as “A”, and the SAAL is used for training. The experimental results have further improved, yielding an AUC score of 0.680 and DP\_20 of 0.939. As for HOTC-2024-RedNIR, the performance of the baseline model “B” yields 0.328 and 0.428. Then, our proposed backbone network increased AUC and DP by 0.3\% and 1.2\%. It is worth noting that SAAM has further increased AUC and DP to 0.357 and 0.463, respectively. It demonstrates the effectiveness of $S^2$FB in extracting spectral information, the superiority of SAFM over intra-modality attention, and the ability of SAAM to supplement the shortcomings of existing tracking heads in object localization prediction.
\begin{table}[]
\caption{ablation experimental results on the HOTC-2020 with the baseline model. $\Delta$ indicates the amount of variation.}
\label{tab:1}
\centering
\begin{tabular}{ccccc}
\hline
Methods & AUC & DP\_20 & $\Delta$(AUC) & $\Delta$(DP\_20) \\ \hline
B        & 0.661  & 0.924  & -       & - \\
S        & 0.673  & 0.932  & +0.012  & +0.008 \\
S+F      & 0.675  & 0.934  & +0.014  & +0.010  \\
S+F+A    & 0.680  & 0.939  & +0.019  & +0.015  \\ \hline
\end{tabular}%
\end{table}

\begin{table}[]
\caption{ablation experimental results on the HOTC-2024-RedNIR with the baseline model. $\Delta$ indicates the amount of variation.}
\label{tab:hot24}
\centering
\begin{tabular}{ccccc}
\hline
Methods & AUC & DP\_20 & $\Delta$(AUC) & $\Delta$(DP\_20) \\ \hline
B        & 0.328  & 0.428  & -       & - \\
S        & 0.331  & 0.440  & +0.003  & +0.012 \\
S+F      & 0.338  & 0.448  & +0.010  & +0.020  \\
S+F+A    & 0.357  & 0.463  & +0.029  & +0.035  \\ \hline
\end{tabular}%
\end{table}

\begin{table}[]
\caption{ablation experimental results of different backbone settings. $\Delta$ indicates the amount of variation compared with the results of the first line.}
\label{tab:2}
\centering
\begin{tabular}{ccccc}
\hline
Methods & AUC & DP\_20 & $\Delta$(AUC) & $\Delta$(DP\_20) \\ \hline
without res       & 0.647  & 0.896  & -  & - \\ 
res            & 0.673  & 0.932  & +0.026  & +0.036 \\ 
spatial        & 0.470  & 0.786  & -0.177  & -0.110 \\ 
spectral       & 0.658  & 0.913  & +0.110  & +0.017 \\ 
both           & 0.675  & 0.934  & +0.028  & +0.038 \\ \hline
\end{tabular}%
\end{table}

\begin{table}[]
\caption{ablation experimental results of different fusion methods. $\Delta$ indicates the amount of variation compared with the results of the first line.}
\label{tab:3}
\centering
\begin{tabular}{ccccc}
\hline
Methods    & AUC & DP\_20 & $\Delta$(AUC) & $\Delta$(DP\_20) \\ \hline
SAFM       & 0.675  & 0.934  & -       & -       \\ 
ADD        & 0.636  & 0.888  & -0.039       & -0.046       \\ 
CONCAT     & 0.662  & 0.926  & -0.013       & -0.008       \\ 
SE-Net \cite{hu2018squeeze}         & 0.665  & 0.930  & -0.010       & -0.004       \\ 
SaE-Net \cite{narayanan2023senetv2}      & 0.670  & 0.927  & -0.005   & -0.007       \\ 
ECA-Net \cite{wang2020eca}    & 0.673  & 0.932  & -0.002       & -0.002       \\ \hline
\end{tabular}%
\end{table}

\begin{table}[]
\caption{ablation experimental results of different prediction strategies. $\Delta$ indicates the amount of variation compared with the results of the first line.}
\label{tab:4}
\centering
\begin{tabular}{ccccc}
\hline
Methods & AUC & DP\_20 & $\Delta$(AUC) & $\Delta$(DP\_20) \\ \hline
weighted prediction  & 0.680 & 0.939 & - & -  \\
average prediction   & 0.639 & 0.915 & -0.041 & -0.024  \\ 
HS prediction        & 0.230 & 0.510 & -0.450 & -0.429  \\
RGB prediction       & 0.623 & 0.871 & -0.057 & -0.068   \\ \hline
\end{tabular}%
\end{table}

\textit{Evaluation of Different Backbone Settings.} To assess the impact of varying backbone architectures, we conduct experiments with the individual spatial and spectral branches of the $S^2$FB and remove the residual components. The experimental results are listed in Table \ref{tab:2}. We first verify the effectiveness of the residual structure, and the experimental results achieve 0.673 of the AUC score and 0.932 of DP\_20, which shows the importance of the residual structure. Next, we explore the branches independently for feature extraction. The spatial branch, solely based on depthwise separable convolutions, lacks comprehensive channel information, resulting in a lower AUC score and DP\_20 of 0.470 and 0.786, respectively. Meanwhile, the spectral branch alone achieved an AUC score of 0.658 and a DP\_20 of 0.913, illustrating that while effective, the performance of each branch is not as robust as when combined. When conducting experiments with the complete $S^2$FB structure, the model performance achieved an AUC score of 0.675 and DP\_20 of 0.934, respectively. It demonstrates that the integrated approach of $S^2$FB can capture both spatial and spectral information and offer a more comprehensive feature representation.

\textit{Evaluation of Different Fusion Methods.} The feature fusion is a key aspect of cross-modality tracking. To show the effectiveness of our proposed SAFM, we conduct experiments with various fusion methods, including element-wise addition (ADD), concatenation (CONCAT), and state-of-the-art spectral attention methods, including SE-Net \cite{hu2018squeeze}, SaE-Net \cite{narayanan2023senetv2}, and ECA-Net \cite{wang2020eca}. As presented in Table \ref{tab:3}, the simplest methods, addition and concatenation, without any attention mechanisms, exhibit the least impressive performance. The current spectral attention mechanisms focus on enhancing within-modality features and fall short in harnessing cross-modality feature interactions. Consequently, the most effective result is offered by ECA-Net, yielding the AUC score and DP\_20 of 0.673 and 0.932, respectively. However, these scores are still 0.2\% lower than those achieved by SAFM. Such a comparison shows the advantages of SAFM in the context of feature fusion, confirming that our module not only refines intra-modality representation but also effectively captures and leverages the correlations across modalities.

\textit{Evaluation of Different Prediction Strategies.} In this section, we evaluate the performance of our weighted prediction approach. Drawing upon the “S” model configuration, we compare four different prediction strategies: using the HS branch alone, using the RGB branch alone, averaging the predictions from both, and applying a weighted prediction method. From Table \ref{tab:4}, it can be seen that using the HS branch to predict only achieves an AUC score of 0.230 and DP\_20 of 0.510. This was expected, as the $S^2$FB is initially trained from scratch on the HOTC-2020 dataset, making it challenging for the HS branch alone to develop robust feature extraction capabilities. Then, the result of the RGB branch is 0.623 of the AUC score and DP\_20 of 0.871. Due to the parameter freezing of the RGB branch, only the knowledge of the RGB domain is used for false-color image tracking, resulting in poor tracking performance. For average prediction, although the prediction results of both modalities are comprehensively considered, the tracking performance reached an AUC score of 0.639 and DP\_20 of 0.915. However, in different scenarios, the contributions of HS and RGB modality to the results should be different. So, we used the weighted prediction method to adjust the weights of the two modality predictions adaptively, and the final results are the AUC score of 0.680 and DP\_20 of 0.939. This further shows the effectiveness of the weighted prediction method.

\textit{Evaluation of Different Loss Weights.} To explain the impact of the weights of the SAAL on the network's performance, we conduct experiments with various weights $\gamma$ of SAAL, including 0.1, 0.5, 1.0, and 1.5, under the settings of $\alpha=1$ and $\beta=2$. The experimental results listed in Table \ref{tab:saal} indicate that setting the weights $\gamma$ to 1.0 yields the best tracking performance. Hence, $\gamma$ is set to 1.0 in our experiments.

\begin{table}[]
\caption{ablation experimental results of different weights of SAAL. $\Delta$ indicates the amount of variation compared with the results of the first line.}
\label{tab:saal}
\centering
\begin{tabular}{ccccc}
\hline
Methods & AUC & DP\_20 & $\Delta$(AUC) & $\Delta$(DP\_20) \\ \hline
0.1  & 0.330 & 0.431 & - & -  \\
0.5   & 0.325 & 0.433 & -0.005 & +0.002  \\ 
1.0        & 0.357 & 0.463 & +0.027 & +0.032  \\
1.5       & 0.338 & 0.464 & +0.008 & +0.033  \\ \hline
\end{tabular}%
\end{table}

\section{Conclusion}

In this article, we proposed a spatial-spectral fusion network with spectral angle awareness (SSF-Net) for hyperspectral object tracking. We first designed a spatial-spectral feature backbone $S^2$FB to extract more robust spatial and spectral features. Then, a spectral attention fusion module (SAFM) was proposed to incorporate the visual information from the RGB modality into the HS modality to obtain robust HS fusion features. Finally, a spectral angle awareness module (SAAM) was designed to predict the more accurate position of objects. With the proposed spectral angle awareness loss (SAAL), the tracker can be trained in an end-to-end way. Experimental results on the HOTC-2020, HOTC-2024, and BihoT dataset showed the effectiveness of our proposed methods and its superiority over the state-of-the-arts.

\ifCLASSOPTIONcaptionsoff
  \newpage
\fi

\bibliographystyle{IEEEtran}
\bibliography{IEEEabrv,ref}

\end{document}